\documentclass[letterpaper, 10 pt, conference]{ieeeconf}  

\IEEEoverridecommandlockouts                              

\usepackage{graphics} 
\usepackage{epsfig} 
\usepackage{times} 
\usepackage{amsmath} 
\usepackage{amssymb}  
\usepackage{color} 
\usepackage{booktabs}
\usepackage{subcaption}
\usepackage[final]{changes}
\usepackage{algorithm}
\usepackage{algorithmic}
\usepackage{calc}
\usepackage{multirow}
\usepackage[hidelinks]{hyperref}

\newtheorem{lemma}{Lemma}

\title{\LARGE \bf
Floating-Base Deep Lagrangian Networks
}

\author{
  Lucas Schulze$^{1,6}$,
  Juliano Decico Negri$^{2}$,
  Victor Barasuol$^{3}$,
  Vivian Suzano Medeiros$^{2}$,\\
  Marcelo Becker$^{2}$,
  Jan Peters$^{1,4,5,6}$,
  Oleg Arenz$^{1,6}$
\thanks{
This work was funded by the German Research Foundation (DFG) - Project number PE 2315/18-1, and by the German Federal Ministry of Research, Technology and Space (BMFTR) - Project number 01IS23057B, and as part of the Robotics Institute Germany (RIG). This project has been supported by a hardware donation by NVIDIA through the Academic Grant Program.
\newline$^{1}$Department of Computer Science, Technical University of Darmstadt, Germany.
$^{2}$Mobile Robotics Group, S\~ao Carlos School of
Engineering, University of S\~ao Paulo (EESC-USP), Brazil.
$^{3}$Dynamic Legged Systems Lab, Istituto Italiano di
Tecnologia (IIT).
$^{4}$Hessian.AI.
$^{5}$German Research Center for AI (DFKI), Research Department: Systems AI for Robot Learning.
$^{6}$Robotics Institute Germany (RIG).
\newline Corresponding author: {\tt\small lucas.schulze@robot-learning.de}
}
}

\newcommand{\Real}{\mathbb{R}}
\newcommand{\inRealTri}{\in \Real^{3}}
\newcommand{\inRealTriTri}{\in \Real^{3\times3}}
\newcommand{\inRealSix}{\in \Real^{6}}
\newcommand{\Transp}{^\mathrm{\scriptscriptstyle T}}

\newcommand{\EyeTr}{\mathbf{1}_3}
\newcommand{\ZeroMat}{\mathbf{0}}
\newcommand{\skewsymSymbol}{\mathbf{S}}
\newcommand{\skewsym}[1]{\skewsymSymbol(#1)}
\newcommand{\rotMat}{\mathbf{R}}

\newcommand{\SEThree}{\mathrm{SE}(3)}
\newcommand{\SOThree}{\mathrm{SO}(3)}
\newcommand{\NegSciExp}[1]{\mathrm{e}{\text{-}#1}}
\newcommand{\PosSciExp}[1]{\mathrm{e}{#1}}
\newcommand{\softplus}{\mathrm{softplus}}

\newcommand{\Lag}{\mathcal{L}}
\newcommand{\EKin}{K}
\newcommand{\EPot}{P}
\newcommand{\inertiaMat}{\mathbf{H}}
\newcommand{\triChol}{\mathbf{C}}
\newcommand{\inertiaTri}{\mathbf{L}}
\newcommand{\LinearFactor}[1]{\inertiaTri_{\mathrm{L}#1}}
\newcommand{\LinearCoupFactor}[1]{\mathbf{L}_{#1\mathrm{L}}}
\newcommand{\RotFactor}[1]{\inertiaTri_{\mathrm{R}#1}}
\newcommand{\RotCoupFactor}[1]{\mathbf{L}_{#1\mathrm{R}}}
\newcommand{\TreeFactor}[1]{\inertiaTri_{#1}}

\newcommand{\FDelanTotalMassParam}{\theta_\mathrm{m}}
\newcommand{\FDelanTotalMass}{\mathrm{m}}
\newcommand{\FDelanTotalMassHat}{\hat{\mathrm{m}}}
\newcommand{\FDelanSkewLRVec}{\mathbf{h}}
\newcommand{\FDelanSkewLRMat}{\skewsymSymbol_\FDelanSkewLRVec}
\newcommand{\FDelanInertiaRotCovFactor}{\mathbf{L}_\Sigma}
\newcommand{\FDelanInertiaRotCov}{\boldsymbol{\Sigma}_\mathrm{R}}
\newcommand{\ProdRotFactor}{\mathbf{D}}
\newcommand{\ProdRotFactorHat}{\hat{\ProdRotFactor}}

\newcommand{\BodyDensityCovar}{\boldsymbol{\Sigma}_\mathrm{b}}

\newcommand{\LinearKinRotFactor}{\mathbf{U}}
\newcommand{\LinearKinFactor}{\mathbf{K}}
\newcommand{\RotKinFactor}{\mathbf{W}}
\newcommand{\ProdLinearFactor}{\mathbf{T}}

\newcommand{\epsilonInertiaTri}{\epsilon_\mathrm{L}}
\newcommand{\epsilonMass}{\epsilon_\FDelanTotalMass}
\newcommand{\epsilonProdRot}{\epsilon_\mathrm{D}}

\newcommand{\ShiftTriIneq}{\beta}
\newcommand{\EigMaxU}{\boldsymbol{\lambda}_{\mathbf{U}}}
\newcommand{\EigMinD}{\mu_{\mathbf{D}}}
\newcommand{\weightU}{w_{\mathbf{U}}}
\newcommand{\weightD}{w_{\mathbf{D}}}

\newcommand{\InertiaNNParam}{\boldsymbol{\phi}}
\newcommand{\NNNorm}{\mathbf{W}_{\jointTau}}

\newcommand{\relMetric}{r\text{NMSE}}

\newcommand{\basePosWF}{\bodyPos_\mathrm{U}^\WF}

\newcommand{\NNRot}{\mathbf{NN}_\mathrm{R}}
\newcommand{\NNKinChain}{\mathbf{NN}}

\newcommand{\WF}{\mathbf{W}}

\newcommand{\TransH}{\mathbf{T}_\inertiaMat}

\newcommand{\jointTau}{\boldsymbol{\tau}}
\newcommand{\jointPos}{\mathbf{q}}

\newcommand{\jointTauRead}{\boldsymbol{\tau}_\mathrm{q}}

\newcommand{\genTau}{\boldsymbol{\tau}_\nu}
\newcommand{\genPos}{\boldsymbol{\nu}}
\newcommand{\genVel}{\dot{\boldsymbol{\nu}}}
\newcommand{\genAcc}{\ddot{\boldsymbol{\nu}}}

\newcommand{\FBSpatialInertia}{\inertiaMat_\mathbf{B}}

\newcommand{\FBInertiaRot}{\mathbf{I}_\mathbf{B}}

\newcommand{\jointFB}{\jointPos_\mathbf{B}}

\newcommand{\TransOmegaToEuler}{\mathbf{W}_{\eta}}
\newcommand{\CoMposang}{\boldsymbol{\Theta}}

\newcommand{\gravityArray}{\mathbf{g}}

\newcommand{\Ncontact}{{\mathrm{n}_\mathrm{c}}}
\newcommand{\JacContact}{\mathbf{J}_\mathrm{c}}
\newcommand{\ForceContact}{\mathbf{f}_\mathrm{c}}

\newcommand{\EoMC}{\mathbf{c}}
\newcommand{\EoMG}{\mathbf{g}}

\newcommand{\NJoints}{{\mathrm{n}_\mathrm{q}}}
\newcommand{\NKinChain}{{\mathrm{n}_\mathrm{k}}}
\newcommand{\Nbodies}{{\mathrm{n}_\mathrm{b}}}

\newcommand{\bodyPos}{\mathbf{r}}

\newcommand{\bodyMass}{\mathrm{m}}
\newcommand{\bodyCoM}{\mathbf{r}_\mathrm{c}}
\newcommand{\bodyInertia}{\mathbf{I}}
\newcommand{\bodyEigInertia}{\mathrm{I}}
\newcommand{\bodySpatialInertia}{\mathbf{M}}

\newcommand{\bodyAngVel}{\boldsymbol{\omega}}

\begin{document}

\maketitle
\thispagestyle{empty}
\pagestyle{empty}

\begin{abstract}
Grey-box methods for system identification combine deep learning with physics-informed constraints, capturing complex dependencies while improving out-of-distribution generalization. Despite the growing importance of floating-base systems such as humanoids and quadrupeds, current grey-box models ignore their specific physical constraints. For instance, the inertia matrix is not only positive definite but also exhibits branch-induced sparsity and input independence. Moreover, the 6×6 composite spatial inertia of the floating base inherits properties of single-rigid-body inertia matrices. As we show, this includes the triangle inequality on the eigenvalues of the composite rotational inertia. To address the lack of physical consistency in deep learning models of floating-base systems, we introduce a parameterization of inertia matrices that satisfies all these constraints. Inspired by Deep Lagrangian Networks (DeLaN), we train neural networks to predict physically plausible inertia matrices that minimize inverse dynamics error under Lagrangian mechanics. For evaluation, we collected and released a dataset on multiple quadrupeds and humanoids. In these experiments, our Floating-Base Deep Lagrangian Networks (FeLaN) achieve better overall performance on both simulated and real robots, while providing greater physical interpretability.

\end{abstract}

\section{INTRODUCTION}
Accurate dynamic models are essential for simulation, control, and state estimation in robotics.
Classical system identification (SysID) constructs a white-box model based on physical principles and estimates its parameters from data. This approach requires prior knowledge of the system, but generalizes well as long as the model assumptions hold.
In contrast, black-box methods approximate the dynamics directly from data without any prior knowledge. Although flexible, they often generalize poorly outside the training domain and lack interpretability.
Grey-box methods combine data-driven models with physical structure. For example, Hamiltonian Neural Networks (HNN)~\cite{hnn_neurips} and Deep Lagrangian Networks (DeLaN)~\cite{delan_lutter2018} embed Hamiltonian and Lagrangian mechanics into deep learning architectures, respectively. 
By enforcing physical properties such as conservation laws, these formulations improve interpretability, generalization, and sample efficiency, and have demonstrated success on fixed-base \cite{delan_4ec} and single
rigid-body~\cite{neural_ode_lie_control, Altawaitan2023HamiltonianDL} robots. Moreover, learned function approximators can yield physically consistent models even with partially or fully unobserved states~\cite{toth2020hamiltoniangenerativenetworks}, 
and can be combined with latent representations~\cite{cadelac_iros25}.
%
\begin{figure}[t]
    \includegraphics{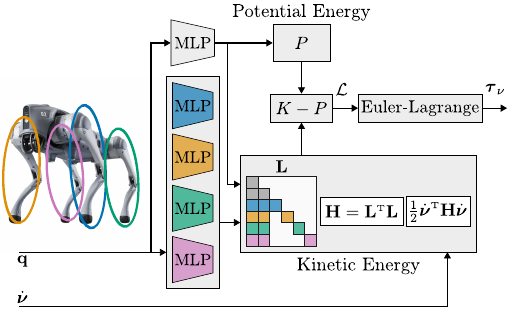}
    \caption{Floating-Base Deep Lagrangian Networks (FeLaN) for Unitree Go2 quadruped.}
    \label{fig:felan_architecture}
\end{figure}

However, naively applying these methods to floating-base systems, such as quadrupeds and humanoids, neglects several physical constraints which may limit the generalization of the learned model. These robots have multiple kinematic chains, leading to a sparse inertia matrix with blocks dependent on specific joint inputs. The composite spatial inertia matrix~\cite{rbd_featherstone2008}, which maps floating-base accelerations to the total wrench applied at the base, must respect a triangle inequality~\cite{full_physical_traversaro}. Even for DeLaN, which predicts the Cholesky decomposition of the inertia matrix, imposing these additional constraints is challenging: sparsity and input-independencies are not preserved in the Cholesky matrix~\cite{efficient_factorization_featherstone2005} and enforcing the triangle inequality is non-trivial. Hence, grey-box models for floating-base robots remain limited to simplified dynamics~\cite{lnn_inf_horizon_quadruped, symbolic_reduced_quadruped_jumping}.

In this work, we introduce a novel parameterization of the inertia matrix of floating-base systems based on the reordered Cholesky factorization~\cite{efficient_factorization_featherstone2005}. Our parameterization ensures positive definiteness; branch-induced sparsity and input-independencies; the triangle inequality of the composite spatial inertia; and the stationarity of the system total mass.
Based on DeLaN, we propose Floating-Base Deep Lagrangian Networks (FeLaN), a novel grey-box method for identifying the dynamics of floating-base systems, depicted in Fig.~\ref{fig:felan_architecture}. We validate our method in several simulation environments and on multiple real quadrupeds and humanoids: Unitree Go2~\cite{UnitreeGo2}, Boston Dynamics Spot~\cite{BostonDynamicsSpot}, Pal Robotics Talos~\cite{Talos}, and HyQReal2, the updated version of HyQReal~\cite{hyqreal19}.

To summarize, our main contributions are as follows:
\begin{itemize}
    \item We extend the full physical consistency condition of single rigid body to the composite spatial inertia.
    \item We present a novel parametrization of the inertia matrix for floating-base systems that preserves branch-induced sparsity and ensures full physical consistency of the composite spatial inertia.
    \item Inspired by DeLaN, we introduce a deep learning architecture for physically consistent system identification of floating-base robots, combining Lagrangian mechanics with our proposed parametrization.
    \item We perform an extensive comparison of our approach against multiple baselines on both simulated and real robot data.
    %
\end{itemize}

The remainder of this paper is organized as follows. Section~\ref{sec:related_work} reviews related work on system identification of floating-base systems. Section~\ref{sec:preliminaries} briefly reviews Lagrangian mechanics and DeLaN. We then present our proposed parametrization of the whole-body inertia matrix in Sec.~\ref{sec:physically_param_inertia_mat} and introduce the deep learning architecture built on this parametrization in Sec.~\ref{sec:floating_base_delan}. Section~\ref{sec:exp} describes the compared baselines and the evaluated experiments. Finally, Section~\ref{sec:conclusion} summarizes our findings and discusses future work.

\section{Related Work}
\label{sec:related_work}
Classical SysID techniques for rigid-body systems usually estimate the kinematic parameters (e.g., link lengths, joint axes) separately from the dynamic parameters (e.g., mass, center of mass, inertia), either by prior estimation or by assuming they are given~\cite{wensing_robot_model_identification}. 
This separation is motivated by the linearity of the equations of motion in the inertial parameters, enabling estimation via linear least squares under persistent excitation~\cite{optimal_excitation}. An alternative approach is to employ differentiable simulators to directly identify all system parameters through stochastic gradient descent~\cite{lutter_diff_sim, pmlr-v120-sutanto20a}.
%
%
For floating-base systems, prior work has explored different sensing modalities, including joint torques~\cite{Khorshidi_physical_no_forces}, joint torques with force sensors~\cite{human_dyn_torque_force}, and force plates~\cite{identif_inertial_legged}.

Conversely, black-box methods employ nonparametric models to learn either the full system dynamics or the residual with respect to a nominal model. For floating-base systems, used methods include Gaussian process regression~\cite{contact_invariant_residual_gpr, anymal_bayesian_mpc}; and neural networks~\cite{neuralsim_nn_residual, nagabandi2017neuralnetworkdynamicsmodelbased}. However, these methods usually model only the forward or inverse dynamics, not holding physical properties.

Integrating physics into deep learning demonstrated a useful inductive bias, especially in small data regimes~\cite{watson2024_survey}. 
For floating-base robots, \cite{lnn_inf_horizon_quadruped} used DeLaN to estimate a simplified formulation of a quadruped's inertia matrix for model-based locomotion. However, only positive definiteness of the inertia is enforced, a necessary but not sufficient condition for full physical consistency, as we prove in Sec.~\ref{subsec:floating_base_inertia}.
%

As the generalized coordinates of a floating-base system include the base pose in $\SEThree$, one could decide to represent the robot's configuration by the pose of each body in $\SEThree$ instead of joint coordinates. For example, \cite{finzi_hnn_exp_constraint, hnn_se3, lie_group_variational_integrator_newtorks} follow this approach, where the inertia matrix simplifies to a block-diagonal concatenation of each body’s spatial inertia. While this formulation can ensure full physical consistency, it introduces redundant coordinates and requires knowledge of the system kinematics, as the body poses are unobserved states. In addition, the reported applications were restricted to systems with a few number of degrees of freedom, e.g., planar pendulums, quadrotors.

Noether’s theorem connects symmetries and conservation laws, motivating their use as an inductive bias~\cite{watson2024_survey}. For instance, \cite{ordonez2025morphosymm} and \cite{morpho_sym_gnn} add symmetry constraints for ground reaction force estimation, demonstrating significant improvements over black-box baselines. However, these methods depend on prior knowledge of the robot’s symmetry groups, which is not always available or generalizable to any robot.

Floating-base systems may have multiple kinematic branches only indirectly coupled through the base, yielding partial functional decoupling.
To exploit this property, \cite{lee2022sampleefficientdynamicslearning} learns an inverse dynamics function for each leg, including the base position expressed in the leg frame, which requires kinematic information.
In fact, the partial functional decoupling is truly described by the branch-induced sparsity of the inertia matrix~\cite{rbd_featherstone2008}, which was first leveraged by \cite{efficient_factorization_featherstone2005} for efficient inertia matrix factorization.

Unlike structural constraints, Sparse Identification of Nonlinear Dynamics (SINDy)~\cite{sindy} promotes sparsity through optimization. \cite{symbolic_reduced_quadruped_jumping} combines linear autoencoders with SINDy to learn a reduced-order quadruped model. While this enforces sparsity, it fails to capture branch-induced sparsity and to guarantee the composite spatial inertia properties.


\section{Preliminaries}
\label{sec:preliminaries}

\subsection{Notation and Definitions}
The $\mathrm{n}\times\mathrm{n}$ identity matrix is denoted as $\mathbf{1}_\mathrm{n}$. $\skewsym{\mathbf{a}}$ and $\mathbf{S}_\mathbf{a}$  denote the skew-symmetric operator over the vector $\mathbf{a}$.
$\mathbf{A} \succeq 0$ and $\mathbf{A} \succ 0$ represents that the symmetric matrix $\mathbf{A}$ is positive semidefinite and definite, respectively. The superscript ${}^\WF$ denotes a variable in the world frame, whereas all other variables are expressed in the base frame. The dependence of matrices and vectors on configuration variables is written explicitly at first mention. 

\subsection{Lagrangian Mechanics}


Lagrangian mechanics formulates rigid-body dynamics in terms of the Lagrangian
\begin{equation}
    \label{eq:lag_def}
    \Lag(\genPos, \genVel) = \EKin(\genPos, \genVel) - \EPot(\genPos) = \frac{1}{2}\genVel\Transp\inertiaMat(\genPos)\genVel - \EPot(\genPos)\text{,}
\end{equation}
where $\genPos$ and $\genVel$ are generalized positions and velocities, $\EKin$ and $\EPot$ are the kinetic and potential energies, and $\inertiaMat(\genPos)$ is the positive-definite inertia matrix.

The equations of motion can be derived from the Euler-Lagrange equation with non-conservative forces $\genTau$
\begin{equation}
    \label{eq:eom_lag_derivative}
    \frac{\partial^2 \Lag(\genPos, \genVel)}{\partial^2\genVel}\genAcc + \frac{\partial \Lag(\genPos, \genVel)}{\partial\genPos\partial\genVel}\genVel -\frac{\partial \Lag(\genPos, \genVel)}{\partial\genPos} = \genTau\text{,}
\end{equation}
where $\genTau$ includes all external forces acting on the system, e.g., actuated joint torques, contact forces. Equation~$\eqref{eq:eom_lag_derivative}$ can be written in the following compact form
\begin{equation}
    \label{eq:eom_rigid_body}
    \inertiaMat(\genPos) \genAcc + \EoMC(\genPos,\genVel) + \EoMG(\genPos) = \genTau\text{,}
\end{equation}
where $\EoMC(\genPos,\genVel)$ and $\EoMG(\genPos)$ are the generalized Coriolis and gravitational forces, respectively.







\subsection{Deep Lagrangian Networks}

Based on \eqref{eq:lag_def}, DeLaN~\cite{delan_2021} approximates the joint-space inertia matrix $\inertiaMat$ and $\EPot$ using two networks to estimate the equations of motion \eqref{eq:eom_rigid_body} of a fixed-base robot. To guarantee physical consistency, i.e., $\inertiaMat \succ 0$, DeLaN parameterizes the inertia matrix in terms of its Cholesky decomposition, i.e.
\begin{equation}
    \label{eq:inertia_chol}
    \inertiaMat(\jointPos) = \triChol(\jointPos)\triChol(\jointPos)\Transp\text{,}
\end{equation}
where $\triChol(\jointPos)$ is a lower triangular matrix with positive diagonal elements.






\subsection{Fully Physically-Consistent Spatial Inertia}
The inertia of a rigid body is defined by its mass $\bodyMass_b \in \Real$, the position of the center of mass in the body frame $\bodyCoM \inRealTri$, the first mass moment $\FDelanSkewLRVec_b  = \bodyMass_b\bodyCoM$, and the rotational inertia $\bodyInertia{}_b\inRealTriTri$ defined at the body frame. These parameters define the body spatial inertia~\cite{rbd_featherstone2008}
\begin{equation}
    \label{eq:body_spatial_inertia}
    \bodySpatialInertia_b = \begin{bmatrix}
    \bodyMass_b\EyeTr & \skewsym{\FDelanSkewLRVec_b}\Transp\\
    \skewsym{\FDelanSkewLRVec_b} &
    \bodyInertia{}_b
    \end{bmatrix}\text{.}
\end{equation}

As $\bodyInertia{}_b$ results from the principal moment of inertia and the positivity of the mass density function, \cite{full_physical_traversaro} demonstrated that the eigenvalues of $\bodyInertia{}_b$, i.e., $\bodyEigInertia_x$, $\bodyEigInertia_y$, $\bodyEigInertia_z$, are not only positive, but also satisfies the triangle inequality
\begin{equation}
    \label{eq:tri_ineq}
    \bodyEigInertia_x + \bodyEigInertia_y \geq \bodyEigInertia_z\text{, }\bodyEigInertia_y + \bodyEigInertia_z \geq \bodyEigInertia_x\text{, }\bodyEigInertia_x + \bodyEigInertia_z \geq \bodyEigInertia_y\text{,}
\end{equation}
which is equivalent to
\begin{equation}
    \label{eq:tri_ineq_trace}
    \frac{1}{2}\mathrm{Tr}(\bodyInertia{}_b) \geq \lambda_\mathrm{max}(\bodyInertia{}_b)
\end{equation}
where $\lambda_\mathrm{max}(\bodyInertia{}_b)$ is the largest eigenvalue of $\bodyInertia{}_b$.
%
%
Hence, full physical consistency is only defined for $\bodyMass_b > 0$ and \eqref{eq:tri_ineq} fulfilled. From \eqref{eq:tri_ineq_trace}, $\cite{lmi_wensing}$ observed that the triangle inequality is equivalent to requiring a positive semidefinite covariance $\BodyDensityCovar{}$ of the body's mass distribution, which are related through
\begin{equation}
    \label{eq:inertia_covar_density_body}
    \bodyInertia{}_b = \mathrm{Tr}(\BodyDensityCovar)\EyeTr - \BodyDensityCovar\text{.}
\end{equation}

\section{Physically Consistent Inertia Matrix Parametrization}
\label{sec:physically_param_inertia_mat}



In this section, we investigate the floating-base inertia matrix $\inertiaMat$ and extend full physical consistency to the composite spatial inertia. We then propose an unconstrained parametrization of $\inertiaMat$ that addresses all identified constraints.

\subsection{Inertia Matrix of a Floating-Base system}
\label{subsec:floating_base_inertia}
For floating-base systems, the generalized coordinates account for both base pose $\jointFB \in \SEThree$, and joint position $\jointPos \in \Real^\NJoints$, leading to $\genPos = [\jointFB; \jointPos]$.
The top-left $6\times6$ block matrix of $\inertiaMat$ corresponds to the composite spatial inertia $\FBSpatialInertia(\jointPos)$ of the whole system~\cite{rbd_featherstone2008}. 
As in \eqref{eq:body_spatial_inertia}, $\FBSpatialInertia$ is defined in the base frame from the total mass $\bodyMass$, total first mass moment $\FDelanSkewLRVec(\jointPos)$, and composite rotational inertia $\FBInertiaRot(\jointPos)$ as
\begin{equation}
    \label{eq:floating_base_spatial_inertia}
    \FBSpatialInertia(\jointPos) = \begin{bmatrix}
    \bodyMass\EyeTr & \skewsym{\FDelanSkewLRVec(\jointPos)}\Transp\\
    \skewsym{\FDelanSkewLRVec(\jointPos)} & \FBInertiaRot(\jointPos)\\
    \end{bmatrix}\text{.}
\end{equation}

Unlike the spatial inertia of a single rigid body, both $\FDelanSkewLRVec(\jointPos)$ and $\FBInertiaRot(\jointPos)$ depend on the joint configuration $\jointPos$ due to the transformation from body to base frame, whereas $\bodyMass$ remains constant and results in an isotropic linear inertia. Since $\FBSpatialInertia$ has the same structure as the spatial inertia of a single rigid body~\eqref{eq:body_spatial_inertia}, $\FBSpatialInertia \succ 0$ is a necessary but not sufficient condition for full physical consistency, as established by Lemma~\ref{lemma:composite_inertia_triangle_inequality}.
\begin{lemma}
\label{lemma:composite_inertia_triangle_inequality}
A composite inertia matrix is fully physically consistent only if $\bodyMass > 0$ and $\FBInertiaRot(\jointPos)$ satisfies the triangle inequality.
\end{lemma}
\begin{proof}
Through the parallel axis theorem, the rotational inertia of the $i$th body with respect to the base frame is
\begin{equation}
    \label{eq:proof_axis_theorem}
    \bodyInertia_i(\jointPos) = \rotMat_i(\jointPos)\bodyInertia_b{}_i\rotMat_i(\jointPos)\Transp + \bodyMass_i\skewsym{\bodyPos_i(\jointPos)}\skewsym{\bodyPos_i(\jointPos)}\Transp\text{,}
\end{equation}
where $\rotMat_i(\jointPos) \in \SOThree$ is the rotation from body to base frame, $\bodyPos_i$ is the body position w.r.t. the base.
The parallel axis theorem preserves the triangle inequality, as it corresponds to a change of reference frame~\cite{full_physical_traversaro}.
From \eqref{eq:tri_ineq_trace}, the trace of the composite rotational inertia satisfies a minimum bound:
\begin{equation}
    \label{eq:proof_trace}
    \mathrm{Tr}(\FBInertiaRot(\jointPos)) = \sum^{\Nbodies}_{i=1}\mathrm{Tr}(\bodyInertia{}_i(\jointPos))  \geq \sum^{\Nbodies}_{i=1} 2\lambda_\mathrm{max}(\bodyInertia{}_i(\jointPos))\text{.}
\end{equation}
Yet, $\bodyInertia{}_i \succ 0$ implies in $\sum^{\Nbodies}_{i=1} \lambda_\mathrm{max}(\bodyInertia{}_i(\jointPos)) \geq \lambda_\mathrm{max}(\FBInertiaRot(\jointPos))$, that combined with \eqref{eq:proof_trace}, yields
\begin{equation}
    \label{eq:tri_ineq_trace_lemma}
    \frac{1}{2}\mathrm{Tr}(\FBInertiaRot(\jointPos)) \geq \sum^{\Nbodies}_{i=1} \lambda_\mathrm{max}(\bodyInertia{}_b{}_i(\jointPos)) \geq \lambda_\mathrm{max}(\FBInertiaRot(\jointPos))\text{,}
\end{equation}
proving the triangle inequality.

\end{proof}


Note that by expressing $\FBSpatialInertia$ in the base frame, $\inertiaMat$ depends only on $\jointPos$ rather than on $\genPos$, yielding translational and rotational invariance with respect to the base coordinates.

Due to the branch-induced sparsity, the joint inertia $\inertiaMat_k$ of the $k$th kinematic branch is a function only of the $k$th branch joints $\jointPos_k$, along with the linear and rotational inertial coupling with the robot's base.

\subsection{Branch-Induced Sparsity}

For fixed-base robots with a single kinematic branch, a positive-definite inertia matrix can be estimated from the Cholesky factor \eqref{eq:inertia_chol}~\cite{delan_lutter2018}.
%
%
%
However, as established by Lemma~\ref{lemma:composite_inertia_triangle_inequality}, $\inertiaMat \succ 0$ is a necessary but not a sufficient condition for a physically consistent $\inertiaMat$ of a floating-base system.
%

The main limitation is that the standard Cholesky factorization~\eqref{eq:inertia_chol} typically results in a dense factor $\triChol(\jointPos)$~\cite{efficient_factorization_featherstone2005}, not holding the branched-induced sparsity and not necessarily the structure of $\FBSpatialInertia(\jointPos)$.
%
To exploit the branch-induced sparsity and efficiently factorize a given $\inertiaMat$, \cite{efficient_factorization_featherstone2005} proposes to use a reordered Cholesky factorization
\begin{equation}
    \label{eq:inertia_reorder_chol}
    \inertiaMat(\jointPos) = \inertiaTri(\jointPos)\Transp\inertiaTri(\jointPos)\text{,}
\end{equation}
where $\inertiaTri(\jointPos)$ is a lower triangular matrix as $\triChol(\jointPos)$. 

The reordered factorization in \eqref{eq:inertia_reorder_chol} is equivalent to a standard Cholesky factorization on a permutation of the original matrix. It preserves the same sparsity pattern from $\inertiaMat$ in $\inertiaTri$, while having the same numerical properties, e.g., positive definiteness. The only additional information required is the bodies' connectivity graph, i.e., the kinematic chains. 


For a robot with $\NKinChain$ kinematic branches, note that $\inertiaTri$ has the following structure:
\begin{equation}
    \label{eq:inertia_tri_floating_base}
    \inertiaTri = \begin{bmatrix}
    \LinearFactor{} & \ZeroMat &  \ZeroMat &  \ZeroMat  & \dots & \ZeroMat \\
    \LinearCoupFactor{\mathrm{R}} & \RotFactor{} & \ZeroMat & \ZeroMat & \dots & \ZeroMat \\
    \LinearCoupFactor{1} & \RotCoupFactor{1} & \TreeFactor{1} & \ZeroMat  & \dots & \ZeroMat \\
    \LinearCoupFactor{2} & \RotCoupFactor{2} & \ZeroMat & \TreeFactor{2} &  \dots & \ZeroMat \\
    \vdots & \vdots & \vdots & \vdots & \ddots & \ddots \\
    \LinearCoupFactor{\NKinChain} & \RotCoupFactor{\NKinChain} & \ZeroMat & \ZeroMat & \dots & \TreeFactor{\NKinChain} \\
    \end{bmatrix}
\end{equation}
where $\LinearFactor{}(\jointPos)$, $\LinearCoupFactor{\mathrm{R}}(\jointPos)$, $\RotFactor{}(\jointPos)$ are a function of all joints $\jointPos$; $\LinearCoupFactor{k}(\jointPos_k)$, $\RotCoupFactor{k}(\jointPos_k)$, $\TreeFactor{k}(\jointPos_{k})$ depend only on $\jointPos_k$ of the $k$th kinematic branch. 


Therefore, any matrix $\inertiaTri$ with the sparsity and functional characteristics in \eqref{eq:inertia_tri_floating_base} implies the same properties in $\inertiaMat$ using \eqref{eq:inertia_reorder_chol}.
%
%
%
Unless otherwise stated, $\inertiaTri$ will hereafter denote the one defined in \eqref{eq:inertia_reorder_chol}.

Based on the structure in \eqref{eq:inertia_tri_floating_base}, we parametrize its matrix blocks to fulfill all physical constraints of $\FBSpatialInertia(\jointPos)$. The following subsections detail how these constraints are ensured.

\subsection{Positive Definiteness}
As the reordered Cholesky~\eqref{eq:inertia_reorder_chol} is equivalent to a factorization on a permuted matrix, $\inertiaTri$ with positive diagonal implies $\inertiaMat \succ 0$. More precisely, this can be achieved by ensuring all the diagonal blocks (e.g., $\LinearFactor{}$, $\RotFactor{}$, $\TreeFactor{1}$, $\dots$, $\TreeFactor{\NKinChain}$) to have positive diagonal elements.


\subsection{Triangle Inequality}


The relation \eqref{eq:inertia_covar_density_body} between a semidefinite covariance matrix and the rotational inertia is defined in any arbitrary frame~\cite{lmi_wensing}.
%
Therefore, we expand its concept to the composite spatial inertia using Lemma~\ref{lemma:composite_inertia_triangle_inequality}, relating a configuration dependent covariance matrix $\FDelanInertiaRotCov(\jointPos) \succ 0$ to a fully physically consistent $\FBInertiaRot(\jointPos)$ by
\begin{equation}
    \FBInertiaRot = \mathrm{Tr}(\FDelanInertiaRotCov)\EyeTr - \FDelanInertiaRotCov\text{,}
\end{equation}
%
where $\FDelanInertiaRotCov(\jointPos) = \FDelanInertiaRotCovFactor(\jointPos)\Transp\FDelanInertiaRotCovFactor(\jointPos)$ with $\FDelanInertiaRotCovFactor(\jointPos) \succ 0$.

To apply this parameterization to $\FBInertiaRot(\jointPos)$, we expand \eqref{eq:inertia_reorder_chol} with \eqref{eq:inertia_tri_floating_base} to obtain
\begin{equation}
    \label{eq:tri_ineq_constraint}
     \RotFactor{}\Transp\RotFactor{} + \RotKinFactor\Transp\RotKinFactor = \FBInertiaRot\text{,}
\end{equation}
where $\RotKinFactor(\jointPos) = [\RotCoupFactor{1}(\jointPos_1); \dots; \RotCoupFactor{\NKinChain}(\jointPos_\NKinChain)]$. Hence, we can express $\RotFactor{}$ in terms of $\FDelanInertiaRotCov$ and $\RotKinFactor(\jointPos)$ as
\begin{align}
        \label{eq:tri_ineq_constraintfinal_term}
        \RotFactor{}\Transp\RotFactor{} &= \FBInertiaRot - \RotKinFactor\Transp\RotKinFactor \\
        &=  \mathrm{Tr}(\FDelanInertiaRotCov)\EyeTr - \FDelanInertiaRotCov  - \RotKinFactor\Transp\RotKinFactor = \ProdRotFactor\text{,}\\
        \label{eq:tri_ineq_constraintfinal_term_rc}
        \RotFactor{} &= \mathrm{reorderedChol} (\ProdRotFactor)\text{ with } \ProdRotFactor \succ 0\text{,}
\end{align}
where $\mathrm{reorderedChol}$ is the factorization proposed by \cite{efficient_factorization_featherstone2005}.





\subsection{First mass moment}
Considering the linear and angular coupling due to the first mass moment given by $\skewsym{\FDelanSkewLRVec(\jointPos)}$ in \eqref{eq:floating_base_spatial_inertia}, expanding \eqref{eq:inertia_reorder_chol} defines the following constraint
\begin{equation}
    \label{eq:skew_sym_coup}
    \LinearCoupFactor{\mathrm{R}}\Transp\RotFactor{} + \LinearKinFactor\Transp\RotKinFactor = \FDelanSkewLRMat\Transp\text{,}
\end{equation}
where $\LinearKinFactor(\jointPos) = [\LinearCoupFactor{1}(\jointPos_1); \dots; \LinearCoupFactor{\NKinChain}(\jointPos_\NKinChain)]$.
Therefore, the constraint \eqref{eq:skew_sym_coup} can be fulfilled by expressing $\LinearCoupFactor{\mathrm{R}}(\jointPos)$ in terms of $\RotFactor{}$, $\LinearKinFactor$, $\RotKinFactor$, and the skew-symmetric matrix $\FDelanSkewLRMat$,
\begin{gather}
    \label{eq:skew_sym_final_term}
    \LinearCoupFactor{\mathrm{R}} = ((\FDelanSkewLRMat\Transp - \LinearKinFactor\Transp\RotKinFactor)\RotFactor{}^{-1})\Transp\text{.}
\end{gather}


\subsection{Mass}

Similarly, the following constraint is defined for the linear inertia of the composite spatial inertia
\begin{equation}
    \label{eq:isotropic_constant_m}
    \LinearFactor{}\Transp\LinearFactor{} + \LinearKinRotFactor\Transp\LinearKinRotFactor = \FDelanTotalMass \EyeTr\text{,}
\end{equation}
where $\LinearKinRotFactor = [\LinearCoupFactor{\mathrm{R}}; \LinearKinFactor]$.
Thus, we express $\LinearFactor{}$ in terms of the total mass $\FDelanTotalMass > 0$ and $\LinearKinRotFactor$ as
\begin{align}
        \label{eq:isotropic_constraint_final_term}
        \LinearFactor{}\Transp\LinearFactor{} &= \FDelanTotalMass \EyeTr - \LinearKinRotFactor\Transp\LinearKinRotFactor = \ProdLinearFactor\text{,}\\
        \label{eq:isotropic_constraint_final_term_rc}
        \LinearFactor{} &= \mathrm{reorderedChol} (\ProdLinearFactor)\text{ with } \ProdLinearFactor \succ 0\text{.}
 \end{align}





\subsection{Branched Kinematic Chains}
Branched kinematic chains, i.e., chains with sub-chains branching from one of their bodies, introduce additional branched-induced sparsity. This type of branching is commonly observed in humanoids, where the arms and head branch from the upper torso, and in quadrupeds with spine joints.
Specifically, the factor $\TreeFactor{kj}$ associated to the sub-chain $j$ of a chain $k$ depends only on $\jointPos_{kj}$.
Consequently, the branch-induced sparsity can be fully exploited through appropriate computation of the elements of $\inertiaTri$, while the remainder of the parametrization and its associated physical consistency properties are preserved.

\subsection{Parametrization Overview}
Given $\FDelanTotalMass$, $\FDelanSkewLRVec$, $\FDelanInertiaRotCovFactor$ and $\LinearCoupFactor{k}, \RotCoupFactor{k}, \TreeFactor{k}$ for each $k$ kinematic branch, we compute an $\inertiaTri(\jointPos)$ which leads to a physically consistent $\inertiaMat$ through \eqref{eq:skew_sym_coup}, \eqref{eq:isotropic_constant_m}, and \eqref{eq:isotropic_constraint_final_term_rc}. 
By directly parameterizing the composite spatial inertia and the joint-space inertia, we require fewer parameters than estimating all 16 kinodynamic parameters for each robot body. Table~\ref{tab:method_compare} shows the number of parameters estimated for each evaluated robot.
%
\begin{table}[h]
    \centering
    \caption{Number of parameters per robot.}
    \setlength{\tabcolsep}{3pt} 
    \begin{tabular}{lccc}
        \toprule
        \multirow{2}{*}{} & Go2 / Spot / HyQReal2 & Spot with Arm & Talos \\
                                & ($\NJoints=12$)
                                & ($\NJoints=19$)
                                & ($\NJoints=22$) \\
        \toprule
        $\inertiaMat$ & 324 & 625 & 784\\
        Standard $\triChol$ & 171 &325 & 406\\
        Reordered $\inertiaTri$ & 117 & 187 & 210\\
        Proposed & 106 & 176 & 199\\
        Body Parameters & 208 & 320 & 368\\ 
        \bottomrule
    \end{tabular}
    \label{tab:method_compare}
\end{table}

\section{Floating-Base Deep Lagrangian Networks}
\label{sec:floating_base_delan}
In this section, we present how we combine deep learning with the parametrization proposed in Sec.~\ref{sec:physically_param_inertia_mat}.

\subsection{Inertia Matrix}
Similar to DeLaN~\cite{delan_2021}, we estimate the elements of $\inertiaTri(\jointPos)$ with deep neural networks. Positive definiteness is satisfied by enforcing positive diagonal elements, achieved by applying an activation function, e.g., ReLU, softplus, combined with an offset $\epsilonInertiaTri$.

To enforce input-independencies for each kinematic branch, we define $\NKinChain$ networks $\NNKinChain_k$ to estimate $\LinearCoupFactor{k}$, $\RotCoupFactor{k}$, and $\TreeFactor{k}$ only from the joints $\jointPos_{k}$ of the $k$th branch. Meanwhile, $\RotFactor{}$ and $\FDelanInertiaRotCovFactor$ are estimated by a single network $\NNRot$ that takes all joints $\jointPos$ as input, 
while $\FDelanTotalMass$ is computed by $\FDelanTotalMassParam^2$, where $\FDelanTotalMassParam \in \Real$ is an unconstrained trainable scalar.

Note that \eqref{eq:tri_ineq_constraintfinal_term} requires $\ProdRotFactor \succ 0$, which is not necessarily true only from the network's outputs. Therefore, to ensure a proper factorization, we define $\ProdRotFactorHat$ using a shifted term $\ShiftTriIneq$ as
\begin{align} \label{eq:tri_ineq_constraint_hat} \ShiftTriIneq &= \epsilonProdRot + \softplus(-\EigMinD)\text{,}\\ 
\label{eq:D_from_tri_ineq_constraint_hat} \ProdRotFactorHat &= \ProdRotFactor + \ShiftTriIneq\EyeTr\text{,} 
\end{align}
where $\EigMinD$ is the smallest eigenvalue of $\ProdRotFactor$, and $\epsilonProdRot$ is a small offset.
Analogously, we ensure $\ProdLinearFactor \succ 0$ in  \eqref{eq:isotropic_constraint_final_term_rc} by
\begin{align}
    \label{eq:mass_hat_pos}
    \FDelanTotalMassHat &= \softplus(\FDelanTotalMass - \max(\EigMaxU)) + \epsilonMass +  \max(\EigMaxU)\text{,}\\
    \label{eq:T_from_mass_hat_pos}
    \ProdLinearFactor &= \FDelanTotalMassHat \EyeTr - \LinearKinRotFactor\Transp\LinearKinRotFactor
\end{align}
where $\EigMaxU$ are the eigenvalues of the product $\LinearKinRotFactor\Transp\LinearKinRotFactor$ and $\epsilonMass$ is a small offset. 
Both $\LinearKinRotFactor\Transp\LinearKinRotFactor$ and $\RotKinFactor\Transp\RotKinFactor$ can be interpreted as the contribution of all the kinematic branches to the composite spatial inertia. Therefore, $\epsilonMass$ and $\epsilonProdRot$ can be interpreted as minimal bounds regarding the contribution of the robot's base to the spatial inertia. Furthermore, $\FDelanTotalMassHat$ and $\ProdRotFactorHat$ do not affect any of the physical consistency constraints. Finally, Algorithm~\ref{alg:inertia_mat} summarizes the inertia matrix computation from the networks' outputs and \eqref{eq:skew_sym_final_term}, \eqref{eq:isotropic_constraint_final_term}, \eqref{eq:mass_hat_pos}, \eqref{eq:T_from_mass_hat_pos}.
\begin{algorithm}
\caption{Inertia Matrix Computation}
\begin{algorithmic}[1]
\REQUIRE joint position $\jointPos$, parameters $(\FDelanTotalMassParam$, $\InertiaNNParam_\mathrm{R}, \dots, \InertiaNNParam_\NKinChain)$
%
\STATE $\FDelanSkewLRVec$, $\FDelanInertiaRotCovFactor{} = \NNRot(\jointPos, \InertiaNNParam_\mathrm{R})$, $\FDelanTotalMass = \FDelanTotalMassParam^2$
\FOR{every $k \in \{1, \ldots, \NKinChain\}$}
\STATE $\LinearCoupFactor{k}, \RotCoupFactor{k}, \TreeFactor{k} =  \NNKinChain_k(\jointPos_k, \InertiaNNParam_k)$
\ENDFOR
\STATE $\FDelanInertiaRotCov{} = \FDelanInertiaRotCovFactor\Transp\FDelanInertiaRotCovFactor$
%
%
\STATE $\LinearKinFactor = [\LinearCoupFactor{1}; \dots, \LinearCoupFactor{\NKinChain}]; \RotKinFactor = [\RotCoupFactor{1}; \dots; \RotCoupFactor{\NKinChain}]$
\STATE $\ProdRotFactor = \mathrm{Tr}(\FDelanInertiaRotCov)\EyeTr - \FDelanInertiaRotCov - \RotKinFactor\Transp\RotKinFactor$
\STATE $\ShiftTriIneq = \epsilonProdRot + \softplus(-\EigMinD)$
\STATE $\ProdRotFactorHat = \ProdRotFactor + \ShiftTriIneq\EyeTr$
\STATE $\RotFactor{} = \mathrm{reorderedChol} (\ProdRotFactorHat)$
\STATE $\LinearCoupFactor{\mathrm{R}} = ((\FDelanSkewLRMat - \LinearKinFactor\Transp\RotKinFactor)\RotFactor{}^{-1})\Transp$
\STATE $\LinearKinRotFactor = [\LinearCoupFactor{\mathrm{R}}; \LinearKinFactor]$
%
\STATE $\FDelanTotalMassHat = \softplus(\FDelanTotalMass -  \max(\EigMaxU)) + \epsilonMass +  \max(\EigMaxU)$
\STATE $\ProdLinearFactor = \FDelanTotalMassHat \EyeTr - \LinearKinRotFactor\Transp\LinearKinRotFactor$
\STATE $\LinearFactor{} = \mathrm{reorderedChol} (\ProdLinearFactor)$
\STATE Assemble $\inertiaTri$ in \eqref{eq:inertia_tri_floating_base} with $\LinearFactor{}, \LinearKinRotFactor, \RotFactor{}, \RotKinFactor, \TreeFactor{1}, \dots, \TreeFactor{\NKinChain}$
\STATE $\inertiaMat = \inertiaTri\Transp\inertiaTri$
\RETURN $\FDelanTotalMassHat$, $\FDelanSkewLRVec$, $\inertiaMat $
\end{algorithmic}
\label{alg:inertia_mat}
\end{algorithm}

\subsection{Potential Energy}


Consider the potential energy of a system with $\Nbodies$ bodies,
\begin{align}
    \label{eq:epot_floating}
    \EPot(\jointPos) &= \sum^{\Nbodies}_{i=1}\bodyMass_i\gravityArray\Transp\bodyPos_i^\WF = \sum^{\Nbodies}_{i=1}\bodyMass_i\gravityArray\Transp(\basePosWF + \rotMat\bodyPos_i(\jointPos)) \\
    \label{eq:epot_floating_sums}
    & = \gravityArray\Transp \left( \sum^{\Nbodies}_{i=1}\bodyMass_i \basePosWF
    + \rotMat\sum^{\Nbodies}_{i=1}\bodyMass_i\bodyPos_i(\jointPos) \right)\text{,}
\end{align}
where $\gravityArray \inRealTri$ is the gravity vector, $\basePosWF \inRealTri$ is the robot's base position in the world frame, $\rotMat$ is the rotation matrix from base to world frame. 
%
Note that the first sum yields the total mass $\FDelanTotalMass$, and the last one the first mass moment $\FDelanSkewLRVec$
\begin{equation}
    \label{eq:epot_from_inertia_param}
    \EPot(\jointPos) = \gravityArray\Transp(\FDelanTotalMass \bodyPos_\mathrm{U}^\WF + \rotMat\FDelanSkewLRVec(\jointPos))\text{.}
\end{equation}


Therefore, we compute the potential energy directly from the estimated terms of $\inertiaMat$, without another network.

\subsection{Generalized Coordinates}

For floating-base systems, the orientation is usually described by Euler angles $\CoMposang$ or quaternions, while the angular velocities are used for the generalized velocities. Consequently, the generalized velocities are not the time derivative of the generalized position. To enable automatic differentiation, we convert the angular velocities to Euler angles time derivatives with $\dot{\CoMposang} = \TransOmegaToEuler(\CoMposang)\bodyAngVel$. 
This transformation also requires converting the angular wrenches to generalized torques.
As the potential energy is defined in the world frame, $\inertiaMat$ in \eqref{eq:inertia_reorder_chol} also has to be transformed from base to world frame. 
Therefore, we apply the following transformations to the total external torque and inertia matrix:
\begin{equation}
    \genTau =
    \begin{bmatrix}
        \mathbf{1}_{3} & \ZeroMat & \ZeroMat \\
        \ZeroMat & \TransOmegaToEuler(\CoMposang)\Transp & \ZeroMat \\
        \ZeroMat & \ZeroMat & \mathbf{1}_{\NJoints}
    \end{bmatrix}\jointTau,
    \quad
    \inertiaMat^\WF = \TransH\Transp\inertiaMat\TransH\text{,}
    \label{eq:inertia_mat_wf}
\end{equation}
with
\begin{equation}
    \TransH = \begin{bmatrix}
        \rotMat(\CoMposang)\Transp & \ZeroMat & \ZeroMat \\
        \ZeroMat & \TransOmegaToEuler(\CoMposang)^{-1} & \ZeroMat \\
        \ZeroMat & \ZeroMat & \mathbf{1}_{\NJoints}
    \end{bmatrix}\text{.}
\end{equation}

\subsection{Loss Function}
\begin{table*}[t]
    \centering
    \caption{Inverse dynamics NMSE over 10 seeds for simulated Go2 and Talos.}
    \setlength{\tabcolsep}{5pt} 
    \begin{tabular}{lccccc}
        \toprule
        Method & \multicolumn{2}{c}{Go2} & \multicolumn{2}{c}{Talos} & $\relMetric$ \\
        & Train & Test & Train & Test & \\
        \midrule
        MLP        & $7.5\NegSciExp{2} \pm 1.1\NegSciExp{2}$ & $1.1\NegSciExp{1} \pm 3.1\NegSciExp{2}$ & $1.9\PosSciExp{0} \pm 1.8\NegSciExp{1}$ & $3.9\PosSciExp{0} \pm 4.9\NegSciExp{1}$ & $9.9\NegSciExp{1}$ \\
        DeLaN      & $2.2\NegSciExp{3} \pm 1.4\NegSciExp{3}$ & $4.2\NegSciExp{3} \pm 6.3\NegSciExp{3}$ & $1.7\NegSciExp{1} \pm 1.3\NegSciExp{2}$ & $2.8\NegSciExp{1} \pm 4.7\NegSciExp{2}$ & $4.4\NegSciExp{2}$ \\
        DeLaN-PP   & $4.0\NegSciExp{3} \pm 1.9\NegSciExp{3}$ & $6.9\NegSciExp{3} \pm 6.9\NegSciExp{3}$ & $8.6\NegSciExp{1} \pm 1.4\PosSciExp{0}$ & $1.1\PosSciExp{0} \pm 1.5\PosSciExp{0}$ & $1.7\NegSciExp{1}$ \\
        FeLaN-BS   & $1.2\NegSciExp{3} \pm 5.5\NegSciExp{4}$ & $1.9\NegSciExp{3} \pm 2.8\NegSciExp{3}$ & $5.1\NegSciExp{2} \pm 1.0\NegSciExp{2}$ & $1.2\NegSciExp{1} \pm 1.5\NegSciExp{1}$ & $1.3\NegSciExp{2}$ \\
        FeLaN      & $1.4\NegSciExp{3} \pm 5.5\NegSciExp{4}$ & $1.7\NegSciExp{3} \pm 1.2\NegSciExp{3}$ & $\mathbf{3.3\NegSciExp{2} \pm 1.1\NegSciExp{2}}$ & $\mathbf{6.7\NegSciExp{2} \pm 3.6\NegSciExp{2}}$ & $\mathbf{4.9\NegSciExp{3}}$ \\
        DNEA NS   & $3.2\NegSciExp{2} \pm 5.7\NegSciExp{2}$ & $2.9\NegSciExp{2} \pm 5.0\NegSciExp{2}$ & $3.8\NegSciExp{1} \pm 6.2\NegSciExp{1}$ & $4.7\NegSciExp{1} \pm 6.7\NegSciExp{1}$ & $1.8\NegSciExp{1}$ \\
        DNEA PD   & $1.7\NegSciExp{3} \pm 4.7\NegSciExp{3}$ & $1.4\NegSciExp{3} \pm 3.5\NegSciExp{3}$ & $3.6\NegSciExp{1} \pm 7.0\NegSciExp{1}$ & $4.3\NegSciExp{1} \pm 7.6\NegSciExp{1}$ & $5.0\NegSciExp{2}$ \\
        DNEA Tri   & $9.5\NegSciExp{4} \pm 1.6\NegSciExp{3}$ & $8.0\NegSciExp{4} \pm 1.1\NegSciExp{3}$ & $5.5\NegSciExp{1} \pm 9.8\NegSciExp{1}$ & $6.8\NegSciExp{1} \pm 1.1\PosSciExp{0}$ & $8.0\NegSciExp{2}$ \\
        DNEA  Cov     & $\mathbf{6.8\NegSciExp{4} \pm 4.9\NegSciExp{5}}$ & $6.4\NegSciExp{4} \pm 2.6\NegSciExp{4}$ & $2.5\NegSciExp{1} \pm 6.8\NegSciExp{1}$ & $2.8\NegSciExp{1} \pm 7.1\NegSciExp{1}$ & $2.8\NegSciExp{2}$ \\
        DNEA Log   & $6.9\NegSciExp{4} \pm 1.3\NegSciExp{4}$ & $\mathbf{6.2\NegSciExp{4} \pm 1.1\NegSciExp{4}}$ & $3.0\NegSciExp{1} \pm 8.2\NegSciExp{1}$ & $3.1\NegSciExp{1} \pm 7.6\NegSciExp{1}$ & $3.2\NegSciExp{2}$ \\
        \bottomrule
    \end{tabular}
    \label{tab:sim_results}
\end{table*}
The Lagrangian \eqref{eq:lag_def} cannot be learned directly with standard supervised learning, as the system's energy is not observed. Following \cite{delan_2021}, the energies are learned indirectly by minimizing the error from the estimated torque $f^{-1}(\genPos, \genVel, \genAcc, \InertiaNNParam)$ given by Algorithm~\ref{alg:torque_computation}, by optimizing
\begin{align}
    \InertiaNNParam^* &= \arg \min_{\InertiaNNParam} \left\| \genTau - f^{-1}(\genPos, \genVel, \genAcc, \InertiaNNParam) \right\|^2_{\NNNorm}\label{eq:opt}\\ 
    &+ \weightU{\textstyle \sum}{}^{\scriptscriptstyle 3}_{\scriptscriptstyle i=1}\softplus(\EigMaxU{}_{i} - \FDelanTotalMass) + \weightD\softplus(-\EigMinD)^2\notag
\end{align}
where $\NNNorm$ is the diagonal covariance of $\genTau$ in the training set; $\weightU$ and $\weightD$ weight the auxiliaries terms from \eqref{eq:tri_ineq_constraint_hat} and \eqref{eq:D_from_tri_ineq_constraint_hat}. Note that physical consistency is achieved as a hard constraint due to the structured parametrization, the extra terms only encourage $\FDelanTotalMass$ and $\ProdRotFactor$ to be positive and positive-definite, respectively, which empirically accelerated training during our experiments.
%
We also apply the input layer $\mathbf{T}_\jointPos (\jointPos) = [\cos(\jointPos); \sin(\jointPos)]$ to each network, as in~\cite{delan_2021}. 

\begin{algorithm}
\caption{Inverse Dynamics}
\begin{algorithmic}[1]
\REQUIRE $\genPos$, $\genVel$, $\genAcc$, $\InertiaNNParam$
\STATE $[\bodyPos_\mathrm{U}^\WF; \CoMposang; \jointPos] = \genPos$
    \STATE $\FDelanTotalMassHat\text{, } \FDelanSkewLRVec(\jointPos)\text{, } \inertiaMat(\jointPos) =\text{Algorithm~\ref{alg:inertia_mat}}(\jointPos, \InertiaNNParam)$%
    \STATE $\EPot = $ \eqref{eq:epot_from_inertia_param} with $\FDelanTotalMassHat$, $\FDelanSkewLRVec(\jointPos)$
    \STATE $\inertiaMat^\WF = \TransH\Transp\inertiaMat\TransH$
    \STATE $\Lag =\frac{1}{2}\genVel\Transp\inertiaMat^\WF\genVel - 
    \EPot$
    \RETURN $\genTau = $ \eqref{eq:eom_lag_derivative} with $\Lag$
\end{algorithmic}
\label{alg:torque_computation}
\end{algorithm}

\section{Experiments}
\label{sec:exp}
\begin{figure*}[t]
    \centering
    \begin{minipage}{\textwidth}
        \centering
        \begin{subfigure}{0.26\textwidth}
            \includegraphics{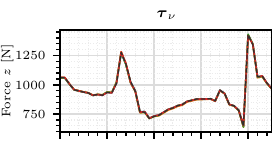}
        \end{subfigure}
        \begin{subfigure}{0.24\textwidth}
            \includegraphics{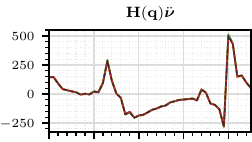}
        \end{subfigure}
        \begin{subfigure}{0.25\textwidth}
            \includegraphics{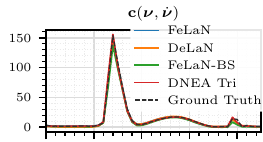}
        \end{subfigure}
        \begin{subfigure}{0.23\textwidth}
            \includegraphics{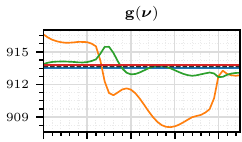}
        \end{subfigure}
    \end{minipage}

    \vspace{0.2em} 

    \begin{minipage}{\textwidth}
        \centering
        \begin{subfigure}{0.26\textwidth}
            \captionsetup{margin={10mm,0mm}}
            \includegraphics{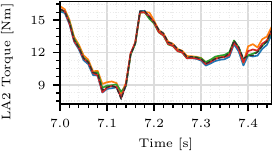}
            \caption{Total Torque}
            \label{subfig:total_torque_sim}
        \end{subfigure}
        \begin{subfigure}{0.24\textwidth}
            \captionsetup{margin={8mm,0mm}}
            \includegraphics{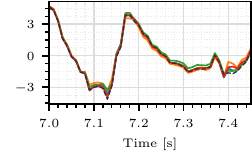}
            \caption{Inertial Torque}
            \label{subfig:inertia_torque_sim}
        \end{subfigure}
        \begin{subfigure}{0.25\textwidth}
            \captionsetup{margin={8mm,0mm}}
            \includegraphics{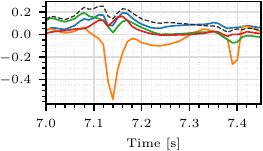}
            \caption{Coriolis Torque}
            \label{subfig:coriolis_torque_sim}
        \end{subfigure}
        \begin{subfigure}{0.23\textwidth}
            \captionsetup{margin={7mm,0mm}}
            \includegraphics{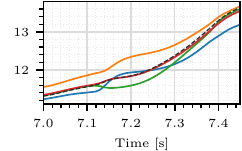}
            \caption{Gravitational Torque}
            \label{subfig:grav_torque_sim}
        \end{subfigure}
    \end{minipage}
    \caption{Components of the estimated force $z$ and left arm second joint (LA2) torque for simulated Talos.}
    \label{fig:sim_torque_plots}
\end{figure*}


We evaluate our proposed architecture for system identification across floating-base robots of different sizes and degrees of freedom on simulation and real-world data.

\subsection{Baselines}

To evaluate the trade-off of physics assumptions and estimation accuracy, we compare our method against different baselines for estimating the inverse dynamics model.
We employ a multilayer perceptron (MLP) with inputs $\CoMposang$, $\jointPos$, $\genVel$, $\genAcc$. Additionally, we use DeLaN~\cite{delan_2021} with one network using $\jointPos$ as input to estimate $\triChol$ in \eqref{eq:inertia_mat_wf}, and a second network that uses the entire $\genPos$ to estimate $\EPot(\genPos)$. To evaluate the proposed potential energy parametrization in \eqref{eq:epot_from_inertia_param}, we implement a modified version of DeLaN, namely DeLaN-PP, which uses a single network for $\inertiaMat(\jointPos)$ and computes $\EPot$ from its estimated terms using \eqref{eq:epot_from_inertia_param}.
We also consider a variant of our proposed network FeLaN, named FeLaN-BS, that incorporates only branch-induced sparsity and positive definiteness. In this variant, $\LinearFactor{}, \LinearCoupFactor{\mathrm{R}},$ and $\RotFactor{}$ are estimated by a single network.
We add the input layer $\mathbf{T}_\jointPos$ to the inputs of all models, except for MLP and the linear coordinates in DeLaN.
We use 2 hidden layers of 32 neurons for MLP, and 2 layers of 16 neurons for the other networks.

Regarding white-box methods, we use Mujoco XLA (MJX)~\cite{todorov2012mujoco} as a differentiable recursive Newton-Euler (DNEA) function, parametrized by 16 kinodynamic parameters per body. In this case, physical consistency is connected to the inertial parameters. Therefore, we evaluate parameterization with different degrees of physical consistency:
\begin{itemize}
    \item DNEA NS~\cite{pmlr-v120-sutanto20a}: unconstrained inertial parameters
    \item DNEA PD~\cite{pmlr-v120-sutanto20a}: positive definiteness via $\bodyInertia{}_b = \triChol_b \triChol_b\Transp$
    \item DNEA Tri~\cite{full_physical_traversaro}: $\eqref{eq:tri_ineq}$ via second moments of mass
    \item DNEA Cov~\cite{lmi_wensing}: parametrize $\BodyDensityCovar = \triChol_b\triChol_b\Transp$ 
    \item DNEA Log~\cite{log_cholesky_wensing}: log-Cholesky factorization of the pseudo-inertia matrix
\end{itemize}


All methods are implemented in JAX~\cite{jax2018github} and trained with stochastic gradient optimization for 2k epochs on quadrupeds and 3k on Talos. We use two hyperparameter sets: one for Talos and one for all the quadrupeds. Results are averaged over 10 seeds. Code and datasets are available online\footnote{\url{https://schulze18.github.io/felan_website}}.

\subsection{Simulated Robots}
For an initial evaluation on robots mainly governed by rigid-body dynamics, we collect data at 100~Hz for Go2 and Talos in MJX. For Talos, we consider only the first four arm joints, while the remaining joints are kept fixed.
%
%
To ensure sufficient excitation and data diversity, we uniformly sample desired velocities and base heights and track them with the model predictive controller~\cite{2025primaldualilqrgpuacceleratedlearning} on both robots.
For Talos’ arms, we provide sine-wave reference trajectories with uniformly sampled amplitudes and frequencies. Each dataset yields a total of 50k samples.
%
%
%
%
As the ground-truth model is available, we compute $\genTau$ using \eqref{eq:eom_rigid_body}. Table~\ref{tab:sim_results} reports the mean and standard deviation over seeds of the torque mean squared error, normalized by $\NNNorm$, for all methods, denoted as NMSE. To compare performance across robots, we define the relative NMSE as $\relMetric_i = (\text{NMSE}_i - \text{NMSE}_{\min})/\text{NMSE}_{\max}$ for each method on the test set, also shown in Tab.~\ref{tab:sim_results}.

\begin{table*}[h]
    \centering
    \caption{Inverse dynamics NMSE over 10 seeds for real robot data.}
    \setlength{\tabcolsep}{4pt}
    \begin{tabular}{lccccccccccc}
        \toprule
        Method & \multicolumn{2}{c}{HyQReal2} & \multicolumn{2}{c}{Spot} 
               & \multicolumn{2}{c}{Spot with Arm} & \multicolumn{2}{c}{Talos} & $\relMetric$ & $\overline{\relMetric}$\\
        & Train & Test & Train & Test & Train & Test & Train & Test && Sim + Real\\
        \midrule
        MLP      & $\mathbf{3.0 \pm 0.4}$ & $\mathbf{4.5 \pm 0.4}$ & $\mathbf{5.9 \pm 0.2}$ & $\mathbf{4.2 \pm 0.3}$ & $\mathbf{7.8 \pm 0.4}$ & $\mathbf{9.3 \pm 0.5}$ & $28.0 \pm 0.1$ & $27.3 \pm 0.1$ & $0.202$ & $0.464$\\
        DeLaN    & $92.5 \pm 3.6$ & $213.1 \pm 104.2$ & ${6.9 \pm 0.1}$ & ${5.1 \pm 0.1}$ & $17.2 \pm 12.9$ & $10.7 \pm 2.2$ & $\mathbf{3.8 \pm 0.2}$ & $\mathbf{5.2 \pm 2.0}$ & $0.307$ & $0.219$\\
        DeLaN-PP & ${6.9 \pm 1.8}$ & ${8.2 \pm 2.4}$ & $8.0 \pm 0.2$ & $5.9 \pm 0.1$ & $11.1 \pm 0.3$ & $9.6 \pm 0.2$ & $7.5 \pm 0.3$ & $7.3 \pm 0.2$ & $\mathbf{0.086}$ & $0.114$\\
        FeLaN-BS & $7.6 \pm 0.4$ & $9.0 \pm 0.6$ & $8.4 \pm 0.1$ & $6.3 \pm 0.1$ & ${10.5 \pm 0.2}$ & ${9.5 \pm 0.3}$ & $8.1 \pm 0.2$ & $7.8 \pm 0.2$ & $0.103$ & $0.073$\\
        FeLaN    & $6.6 \pm 0.1$ & $7.8 \pm 0.2$ & $8.1 \pm 0.1$ & $6.0 \pm 0.1$ & ${10.5 \pm 0.2}$ & $9.6 \pm 0.2$ & $7.5 \pm 1.0$ & $7.2 \pm 0.5$ & $0.090$  & $\mathbf{0.062}$\\
        DNEA NS & $12.8 \pm 0.2$ & $14.5 \pm 0.2$ & $10.1 \pm 0.1$ & $7.6 \pm 0.1$ & $14.4 \pm 0.2$ & $10.9 \pm 0.2$ & $15.9 \pm 0.5$ & $14.8 \pm 0.5$ & $0.248$  & $0.225$\\
        DNEA PD & $12.9 \pm 0.2$ & $14.7 \pm 0.3$ & $10.1 \pm 0.1$ & $7.6 \pm 0.1$ & $14.4 \pm 0.2$ & $10.9 \pm 0.2$ & $17.1 \pm 1.0$ & $16.0 \pm 1.0$ & $0.260$  & $0.190$\\
        DNEA Tri & $13.0 \pm 0.2$ & $14.8 \pm 0.2$ & $10.2 \pm 0.1$ & $7.6 \pm 0.1$ & $14.5 \pm 0.2$ & $11.0 \pm 0.2$ & $17.5 \pm 1.0$ & $16.4 \pm 0.8$ & $0.266$  & $0.204$\\
        DNEA Cov    & $13.0 \pm 0.2$ & $14.8 \pm 0.2$ & $10.1 \pm 0.1$ & $7.6 \pm 0.1$ & $14.5 \pm 0.3$ & $11.0 \pm 0.3$ & $17.5 \pm 1.0$ & $16.4 \pm 0.8$ & $0.266$  & $0.186$\\
        DNEA Log & $13.2 \pm 0.1$ & $15.0 \pm 0.2$ & $10.2 \pm 0.1$ & $7.6 \pm 0.1$ & $14.5 \pm 0.2$ & $11.0 \pm 0.2$ & $17.7 \pm 1.0$ & $16.5 \pm 0.9$ & $0.268$ & $0.189$\\
        \bottomrule
    \end{tabular}
    \label{tab:real_results}
\end{table*}
%
%


%
For Go2, the fully physically consistent DNEAs achieve the best performance, as their consistent parametrization fully captures the system's dynamics. All grey-box methods perform well, with FeLaN and FeLaN-BS outperforming DNEA NS, demonstrating the importance of physical consistency.
For Talos, FeLaN performed best, followed by FeLaN-BS. Although the amplitude errors remain small, the slightly worse performance of DNEA on Talos may be due to contact changes, adding larger spikes in the data when compared to Go2, making the estimation more challenging. DeLaN still performed similarly compared to the other methods, while DeLaN-PP demonstrated sensitivity to different initialization. For both robots, the MLP performed the worst. Accordingly, FeLaN demonstrated the most consistent results, with the lowest $\relMetric$, followed by FeLaN-BS.

Figure~\ref{fig:sim_torque_plots} shows the components in \eqref{eq:eom_rigid_body} for the force $z$ and one torque joint of Talos. All grey-box methods approximate the total value well, see Fig.~\ref{subfig:total_torque_sim}. FeLaN better predicts the individual terms, in particular the stationary gravity component of $z$ in Fig.~\ref{subfig:grav_torque_sim}, due to its proper mass parametrization.

\subsection{Real Robots}
To evaluate all methods on real-world data, we ran experiments on four robots: HyQReal2, Spot, Spot with Arm, and Talos.
As a ground truth model is not available, we estimate $\genTau$, which for a legged robot with $\Ncontact$ contacts is given by
\begin{equation}
    \label{eq:joint_total_contact}
    \genTau = \begin{bmatrix}
        \ZeroMat \\ \jointTauRead
    \end{bmatrix} +  \sum^{\Ncontact}_{i=1} \JacContact\Transp{}_i\ForceContact{}_i
\end{equation}
where $\jointTauRead \in \Real^\NJoints$ are the joint torques; $\JacContact{}_i$ and $\ForceContact{}_i$ are the $i$th contact Jacobian and force. 
All evaluated robots have torque sensors. For contact forces, Talos uses an ankle sensor, HyQReal2 relies on a force estimator based on its dynamical model, and Spot’s forces were measured with external 6D force plates.
Although the estimation of $\genTau$ requires kinematics information, all the methods still learn agnostically to any kinematic parameter.
For HyQReal2 and Talos, trajectories similar to those in simulation were executed, producing 75k and 125k samples, respectively. To keep Spot on the force plates, small velocity commands were sent while trotting in place, resulting in 60k samples. 
For the arm, a position controller tracked sine-wave trajectories with uniformly sampled parameters, yielding 45k samples.

Table~\ref{tab:real_results} reports the NMSE results, including the weighted average $\overline{\relMetric}$ over real and simulated robots. Unlike in simulation, real robot dynamics are affected by actuator and contact dynamics, noise, and other unmodeled effects. Consequently, DNEA methods struggle to accurately estimate the torques due to limited model complexity, being in general outperformed by most of the other methods.
Following the simulation results on Talos, the performance gap of DNEAs is larger for the heavier robots. On HyQReal2 and Talos, 125 kg and 100 kg, respectively, DNEA errors are roughly three times higher than the best. 
Their larger mass implies larger contact forces and thereby larger torques than on Spot. 
This tends to amplify the influence of actuator and contact dynamics that the models do not capture, thereby limiting prediction accuracy.
In contrast, DeLaN and MLP achieve the best robot-specific results. Because these methods impose weaker or no physical constraints, they are more flexible in learning input–output mappings that may not be physically consistent. However, they demonstrated sensitivity to initialization and hyperparameter choices, and do not perform consistently across all robots.



Overall, DeLaN-PP, FeLaN-BS, and FeLaN remain competitive with the best robot-specific models while achieving the best $\relMetric$. 
This supports incorporating physics into deep learning by improving consistency across robots, reducing sensitivity to dataset-specific biases. In particular, it highlights the effectiveness of our potential energy parameterization.
%
%
Considering $\overline{\relMetric}$, FeLaN achieves the best overall performance across simulation and real robots, followed by FeLaN-BS, underscoring the benefit of branch-induced sparsity and full physical consistency of the composite spatial inertia. 

%


\section{Conclusion}
\label{sec:conclusion}

In this work, we introduced a novel physically consistent parametrization of the floating-base inertia matrix, based on a reordered Cholesky factorization, which achieves branch-induced sparsity and ensures full physical consistency of the composite spatial inertia. Building on this parametrization, we proposed FeLaN, a grey-box method for physically consistent system identification of floating-base dynamics. FeLaN learns the whole-body dynamics model using only prior knowledge on the kinematic chains, without requiring any knowledge of dynamics or other kinematic parameters. 
We validated FeLaN on both simulated and real robot data, showing overall better performance than state-of-the-art methods while offering greater interpretability.

As future work, we plan to explore using FeLaN for locomotion and loco-manipulation tasks, either as a learned dynamical model or alongside adaptive control techniques.






\addtolength{\textheight}{-0.1cm}
\bibliographystyle{./bibtex/IEEEtran}
\bibliography{references}

\end{document}